\documentclass[11pt,a4paper]{article}
\usepackage{authblk}
\usepackage[hyperref]{naaclhlt2018}
\usepackage{times}
\usepackage{latexsym}
\usepackage{amssymb}
\usepackage{amsmath}
\usepackage{authblk}
\usepackage{url}
\usepackage{graphicx}
\usepackage{booktabs}
\aclfinalcopy 
\def\aclpaperid{964} 

\title{
Pragmatically Informative Image Captioning\\ with Character-Level Inference
}

\author[ ]{Reuben Cohn-Gordon}
\author[ ]{Noah Goodman}
\author[ ]{Chris Potts}
\affil[ ]{Stanford University}
\affil[ ]{\tt \{reubencg, ngoodman, cgpotts\}@stanford.edu}

\begin{document}
\maketitle

\begin{abstract}

We combine a neural image captioner with a Rational Speech Acts (RSA) model to make a system that is \emph{pragmatically informative}: its objective is to produce captions that are not merely true but also distinguish their inputs from similar images. Previous attempts to combine RSA with neural image captioning require an inference which normalizes over the entire set of possible utterances. This poses a serious problem of efficiency, previously solved by sampling a small subset of possible utterances. We instead solve this problem by implementing a version of RSA which operates at the level of characters (``a'',``b'',``c'', \ldots) during the unrolling of the caption. We find that the utterance-level effect of referential captions can be obtained with only character-level decisions. Finally, we introduce an automatic method for testing the performance of pragmatic speaker models, and show that our model outperforms a non-pragmatic baseline as well as a word-level RSA captioner.

\end{abstract}

\section{Introduction}\label{sec:intro}

The success of automatic image captioning
\cite{farhadi2010every,mitchell2012midge,karpathy2015deep,vinyals2015show}
demonstrates compellingly that end-to-end
statistical models can align visual information with language. However,
high-quality captions are not merely \emph{true}, but also
\emph{pragmatically informative} in the sense that they highlight salient properties and help distinguish their inputs from
similar images. Captioning systems trained on single images struggle
to be pragmatic in this sense, producing either very general or
hyper-specific descriptions.

\begin{figure}
\includegraphics[width=\linewidth, height=4cm]{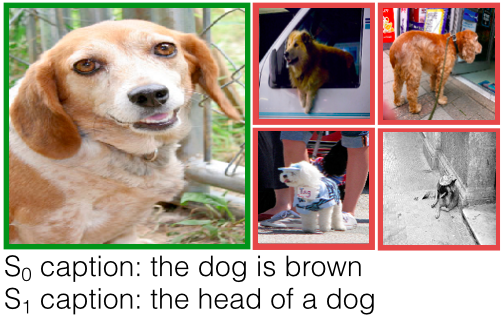}
\caption{Captions generated by literal ($S_0$) and pragmatic ($S_1$) model for the target image (in green) in the presence of multiple distractors (in red).}
\label{fig1}
\end{figure}

In this paper, we present a neural image captioning system\footnote{The code is available at \url{https://github.com/reubenharry/Recurrent-RSA}} that is a
\emph{pragmatic speaker} as defined by the Rational Speech Acts (RSA)
model \cite{Frank:Goodman:2012,goodman2}. Given a set of images, of
which one is the \emph{target}, its objective is to generate a natural
language expression which identifies the target in this
context. For instance, the literal caption in Figure~\ref{fig1} could describe both the target and the top two distractors, whereas the pragmatic caption mentions something that is most salient of the target. Intuitively, the RSA speaker achieves this by
reasoning not only about what is true but also about what it's like to
be a listener in this context trying to identify the target.

This core idea underlies much work in referring expression generation
\cite{Dale:Reiter:1995,Monroe:Potts:2015,andreas2016reasoning,monroe2016learning}
and image captioning \cite{mao2016generation,murphy}, but these
models do not fully confront the fact that the agents must reason
about all possible utterances, which is intractable. We fully address
this problem by implementing RSA at the level of characters rather
than the level of utterances or words: the neural language model emits individual characters,
choosing them to balance pragmatic informativeness with overall
well-formedness. Thus, the agents reason not about full utterances, but rather only about all possible character choices, a very small space. The result is that the information encoded recurrently in the neural model allows us to obtain global pragmatic effects from local  decisions. We show that such character-level RSA speakers are more effective than literal captioning systems at the task of helping a reader identify the target image among close competitors, and outperform word-level RSA captioners in both efficiency and accuracy.

\section{Bayesian Pragmatics for Captioning} \label{rsaintro}

In applying RSA to image captioning, we think of captioning as a
kind of reference game. The \emph{speaker} and \emph{listener} are in a shared
context consisting of a set of images $W$, the speaker is privately
assigned a target image $w^{\ast} \in W$, and the speaker's goal is to
produce a caption that will enable the listener to identify
$w^{\ast}$. $U$ is the set of possible utterances. In its simplest form, the \emph{literal speaker} is a
conditional distribution $S_{0}(u|w)$ assigning equal probability to all true utterances $u\in U$ and $0$ to all others. The pragmatic listener
$L_{0}$ is then defined in terms of this literal agent and a prior $P(w)$ over possible images:
\begin{equation}
L_{0}(w|u) \propto \frac{S_{0}(u|w)*P(w)}{\sum_{w'\in W}S_{0}(u|w')*P(w')}
\end{equation}
The pragmatic speaker $S_{1}$ is then defined in terms of this pragmatic listener, with the addition of a rationality parameter
$\alpha > 0$ governing how much it takes into account the $L_{0}$ 
distribution when choosing utterances. $P(u)$ is here taken to be a uniform distribution over $U$:
\begin{equation}
S_{1}(u|w) \propto \frac{L_{0}(w|u)^{\alpha}*P(u)}{\sum_{u'\in U}L_{0}(w|u')^{\alpha}*P(u')}
\end{equation}
As a result of this back-and-forth, the $S_{1}$ speaker is reasoning not merely about what is true, but rather about a listener reasoning about a literal speaker who reasons about truth.

\begin{figure}

\includegraphics[width=\linewidth, height=4cm]{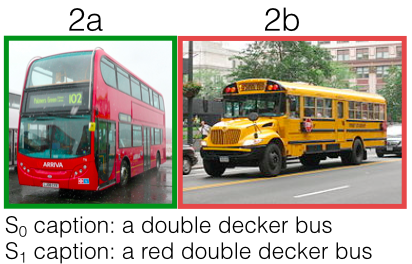}
\caption{Captions for the target image (in green).}
\label{2}
\end{figure}

To illustrate, consider the pair of images 2a and 2b in Figure~\ref{2}. Suppose that $U =\{\emph{bus}, \emph{red bus}\}$. Then the literal speaker $S_{0}$ is equally likely to produce \emph{bus} and \emph{red bus} when the left image 2a is the target. However, $L_{0}$ breaks this symmetry; because \emph{red bus} is false of the right bus, $L_0(\ref{2}\textrm{a}|\mathit{bus}) = \frac{1}{3}$ and $L_0(\ref{2}\textrm{b}|\mathit{bus}) = \frac{2}{3}$. The $S_{1}$ speaker therefore ends up favoring \emph{red bus} when trying to convey 2a, so that $S_1(\emph{red bus}|2\textrm{a}) = \frac{3}{4}$ and $S_1(\mathit{bus}|2\textrm{a}) = \frac{1}{4}$.

\section{Applying Bayesian Pragmatics to a Neural Semantics}

To apply the RSA model to image captioning, we first train a neural model with a CNN-RNN architecture \cite{karpathy2015deep,vinyals2015show}. The trained model can be considered an $S_{0}$-style distribution $P(\emph{caption}|\emph{image})$
on top of which further listeners and speakers can be built. (Unlike the idealized $S_0$ described above, a neural $S_0$ will assign some probability to untrue utterances.)

The main challenge for this application is that the space of utterances (captions) $U$ will be very large for any suitable captioning system, making the calculation of $S_{1}$ intractable due to its normalization over all utterances.
The question, therefore, is how best to approximate this inference. The solution employed by \citet{monroe2016learning} and \citet{andreas2016reasoning} is to sample a small subset of probable utterances from the $S_0$, as an approximate prior upon which exact inference can be performed. While tractable, this approach has the shortcoming of only considering a small part of the true prior, which potentially decreases the extent to which pragmatic reasoning will be able to apply. In particular, if a useful caption never appears in the sampled prior, it cannot appear in the posterior.

\subsection{Step-Wise Inference}

Inspired by the success of the ``emittor-suppressor'' method of \citet{murphy}, we propose an incremental version of RSA. Rather than performing a single inference over utterances, we perform an inference \emph{for each step of the unrolling of the utterance}. 

We use a character-level LSTM, which defines a distribution over characters $P(u|\emph{pc},\emph{image})$, where $\emph{pc}$ (``partial caption'') is a string of characters constituting the caption so far and $u$ is the next character of the caption. This is now our $S_0$: given a partially generated caption and an image, it returns a distribution over which character should next be added to the caption. The advantage of using a character-level LSTM over a word-level one is that $U$ is much smaller for the former (${\approx}30$ vs.~${\approx}20,000$), making the ensuing RSA model much more efficient.

We use this $S_0$ to define an $L_0$ which takes a partial caption and a new character, and returns a distribution over images. The $S_1$, in turn, given a target image $w^{\ast}$, performs an inference over the set of possible characters to determine which is best with respect to the listener choosing $w^{\ast}$.

At timestep $t$ of the unrolling, the listener $L_0$ takes as its prior over images the $L_0$ posterior from timestep $(t-1)$. The idea is that as we proceed with the unrolling, the $L_0$ priors on which image is being referred to may change, which in turn should affect the speaker's actions. For instance, the speaker, having made the listener strongly in favor of the target image, is less compelled to continue being pragmatic. 

\subsection{Model Definition} \label{decision2}

In our incremental RSA, speaker models take both a target image and a partial caption \emph{pc}. Thus, $S_0$ is a neurally trained conditional distribution $S_0^t(u|w,\emph{pc}_t)$, where $t$ is the current timestep of the unrolling and $u$ is a character.

We define the $L_0^{t}$ in terms of the $S_0^{t}$ as follows, where \emph{ip} is a distribution over images representing the $L_0$ prior:

\begin{equation}
L_0^t(w|u,\emph{ip}_t,\emph{pc}_t) \propto 
	S_0^t(u|w,\emph{pc}_t)  * \emph{ip}_t(w)
\end{equation}

\noindent
Given an $S_0^{t}$ and $L_0^{t}$, we define $S_1^{t}$ and $L_1^{t}$ as:
\begin{multline}
S_1^t(u|w,\emph{ip}_t,\emph{pc}_t) \propto \\
	S_{0}^t(u|w,\emph{pc}_t) * 		
    L_{0}^t(w|u,\emph{ip}_t,\emph{pc}_t)^{\alpha}
\end{multline}

\vspace{-20pt}

\begin{multline}
L_1^t(w|u,\emph{ip}_t,\emph{pc}_t) \propto  \\
	L_{0}^t(w|u,\emph{ip}_t,\emph{pc}_t) * 
    S_{0}^t(u|w,\emph{pc}_t)
\end{multline}

\paragraph{Unrolling} To perform greedy unrolling (though in practice we use a beam search) for either $S_0$ or $S_1$, we initialize the state as a partial caption $\emph{pc}_{0}$ consisting of only the start token and a uniform prior over the images $\emph{ip}_{0}$. 
Then, for $t > 0$, we use our incremental speaker model $S_0$ or $S_1$ to generate a distribution over the subsequent character $S^{t}(u | w,\emph{ip}_t,\emph{pc}_t)$, and add the character $u$ with highest probability density to $\emph{pc}_t$, giving us $\emph{pc}_{t+1}$. We then run our listener model $L_1$ on $u$, to obtain a distribution $\emph{ip}_{t+1}=L_{1}^{t}(w | u,\emph{ip}_t,pc_{t})$ over images that the $L_0$ can use at the next timestep.

This incremental approach keeps the inference itself very simple, while placing the complexity of the model in the recurrent nature of the unrolling.\footnote{The move from standard to incremental RSA can be understood as a switching of the order of two operations; instead of unrolling a character-level distribution into a sentence level one and then applying pragmatics, we apply pragmatics and then unroll. This generalizes to any recursive generation of utterances.}  While our $S_0$ is character-level, the same incremental RSA model works for a word-level $S_0$, giving rise to a word-level $S_1$. We compare character and word $S_1$s in section \ref{section:5}.

As well as being incremental, these definitions of $S_1^{t}$ and $L_1^{t}$ differ from the typical RSA described in section \ref{rsaintro} in that $S_1^{t}$ and $L_1^{t}$ draw their priors from $S_{0}^{t}$ and $L_{0}^{t}$ respectively. This generalizes the scheme put forward for $S_1$ by \newcite{andreas2016reasoning}. The motivation is to have Bayesian speakers who are somewhat constrained by the $S_0$ language model. Without this, other methods are needed to achieve English-like captions, as in \newcite{murphy}, where their equivalent of the $S_1$ is combined in a weighted sum with the $S_0$. 

\section{Evaluation}

Qualitatively, Figures~\ref{fig1} and \ref{2} show how the $S_1$ captions are more informative than the $S_0$, as a result of pragmatic considerations. To demonstrate the effectiveness of our method quantitatively, we implement an automatic evaluation. 

\subsection{Automatic Evaluation}

To evaluate the success of $S_1$  as compared to $S_0$, we define a listener $L_{\emph{eval}}(\emph{image}|\emph{caption}) \propto P_{S_0}(\emph{caption}|\emph{image})$, where $P_{S_0}(\emph{caption}|\emph{image})$ is the total probability of $S_0$ incrementally generating $\emph{caption}$ given $\emph{image}$. In other words, $L_{\mathit{eval}}$ uses Bayes' rule to obtain from $S_0$ the posterior probability of each image $w$ given a full caption $u$. 

The neural $S_0$ used in the definition of $L_{\emph{eval}}$ must be trained on separate data to the neural $S_0$ used for the $S_1$ model which produces captions, since otherwise this $S_1$ production model effectively has access to the system evaluating it. As \citet{backprop} note, ``a model might `communicate' better with itself using its own language than with others''. In evaluation, we therefore split the training data in half, with one part for training the $S_0$ used in the caption generation model $S_1$ and one part for training the $S_0$ used in the caption evaluation model $L_{\emph{eval}}$.

We say that the caption succeeds as a referring expression if the target has more probability mass under the distribution $L_{\emph{eval}}(\emph{image}|\emph{caption})$ than any distractor.

\paragraph{Dataset} We train our production and evaluation models on separate sets consisting of regions in the Visual Genome dataset \cite{krishnavisualgenome} and full images in MSCOCO \cite{mscoco}. Both datasets consist of over 100,000 images of common objects and scenes. MSCOCO provides captions for whole images, while Visual Genome provides captions for regions within images.

Our test sets consist of clusters of 10 images. For a given cluster, we set each image in it as the target, in turn. We use two test sets. Test set~1 (TS1) consists of 100 clusters of images, 10 for each of the 10 most common objects in Visual Genome.\footnote{Namely, \emph{man}, \emph{person}, \emph{woman}, \emph{building}, \emph{sign}, \emph{table}, \emph{bus}, \emph{window}, \emph{sky}, and \emph{tree}.}

Test set~2 (TS2) consists of regions in Visual Genome images whose ground truth captions have high word overlap, an indicator that
they are similar. We again select 100 clusters of 10. Both test sets have 1,000 items in total (10 potential target images for each of 100 clusters).

\paragraph{Captioning System} Our neural image captioning system is a CNN-RNN architecture\footnote{\url{https://github.com/yunjey/pytorch-tutorial/tree/master/tutorials/03-advanced/image_captioning}} adapted to use a character-based LSTM for the language model.

\paragraph{Hyperparameters} We use a beam search with width 10 to produce captions, and a rationality parameter of $\alpha=5.0$ for the $S_1$.

\subsection{Results} \label{section:5}

As shown in Table~\ref{fig3}, the character-level $S_1$ obtains higher accuracy (68\% on TS1 and 65.9\% on TS2) than the $S_0$ (48.9\% on TS1 and 47.5\% on TS2), demonstrating that $S_1$ is better than $S_0$ at referring. 

\paragraph{Advantage of Incremental RSA}
We also observe that 66\% percent of the times in which the $S_1$ caption is referentially successful and the $S_0$ caption is not, for a given image, the $S_1$ caption is not one of the top 50 $S_0$ captions, as generated by the beam search unrolling at $S_0$. This means that in these cases the non-incremental RSA method of \newcite{andreas2016reasoning} could not have generated the S$_1$ caption, if these top 50 $S_0$ captions were the support of the prior over utterances.

\begin{table}[t!]
\begin{center}
\setlength{\tabcolsep}{12pt}
\begin{tabular}{l r l}
\toprule 
\bf Model & \bf TS1 & \bf TS2 \\ 
\midrule
Char $S_0$ & $48.9$   & $47.5$ \\
Char $S_1$ & $\mathbf{68.0}$ & $\mathbf{65.9}$ \\
Word $S_0$ & $57.6$   & $53.4$ \\
Word $S_1$ & $60.6$   &  $57.6$ \\
\bottomrule
\end{tabular}
\end{center}
\caption{\label{fig3} Accuracy on both test sets.} 
\end{table}

\paragraph{Comparison to Word-Level RSA} We compare the performance of our character-level model to a word-level model.\footnote{Here, we use greedy unrolling, for reasons of efficiency due to the size of $U$ for the word-level model, and set $\alpha=1.0$ from tuning on validation data. For comparison, we note that greedy character-level $S_1$ achieves an accuracy of 61.2\% on TS1.} This model is incremental in precisely the way defined in section \ref{decision2}, but uses a word-level LSTM so that $u\in U$ are words and $U$ is a vocabulary of English. It is evaluated with an $L_{\mathit{eval}}$ model that also operates on the word level. 

Though the word $S_0$ performs better on both test sets than the character $S_0$, the character $S_1$ outperforms the word $S_1$, demonstrating the advantage of a character-level model for pragmatic behavior. We conjecture that the superiority of the character-level model is the result of the increased number of decisions where pragmatics can be taken into account, but leave further examination for future research.

\paragraph{Variants of the Model} We further explore the effect of two design decisions in the character-level model. First, we consider a variant of $S_1$ which has a prior over utterances determined by an LSTM language model trained on the full set of captions. This achieves an accuracy of 67.2\% on TS1. Second, we consider our standard $S_1$ but with unrolling such that the $L_0$ prior is drawn uniformly at each timestep rather than determined by the $L_0$ posterior at the previous step. This achieves an accuracy of 67.4\% on TS1. This suggests that neither this change of $S_1$ nor $L_0$ priors has a large effect on the performance of the model.

\section{Conclusion}

We show that incremental RSA at the level of characters improves the ability of the neural image captioner to refer to a target image. The incremental approach is key to combining RSA with language models: as utterances become longer, it becomes exponentially slower, for a fixed $n$, to subsample $n$\% of the utterance distribution and \emph{then} perform inference (non-incremental approach). Furthermore, character-level RSA yields better results than word-level RSA and is far more efficient. 

\section*{Acknowledgments}

Many thanks to Hiroto Udagawa and Poorvi Bhargava, who were involved in early versions of this project. This material is based in part upon work supported by the Stanford Data Science Initiative and by the NSF under Grant No.~BCS-1456077. This work is also supported by a Sloan Foundation Research Fellowship to Noah Goodman.

\bibliography{naaclhlt2018}
\bibliographystyle{acl_natbib_alt}

\appendix

\end{document}